\pdfoutput=1

\documentclass[10pt,twocolumn,letterpaper]{article}

\usepackage{cvpr}
\usepackage{times}
\usepackage{graphicx}
\usepackage{amsmath}
\usepackage{amssymb}

\usepackage{comment}

\usepackage{bbold}
\usepackage{stmaryrd}

\graphicspath{{figures/}}

\newcommand{\Eq}[1]{Eq.~(\ref{eq:#1})}
\newcommand{\eq}[1]{\Eq{#1}}

\newcommand{\fig}[1]{Fig.~\ref{fig:#1}}
\newcommand{\tab}[1]{Table~\ref{tab:#1}}

\newcommand{\sect}[1]{Section~\ref{sec:#1}}

\usepackage{booktabs}

\usepackage{xspace}

\cvprfinalcopy 

\def\cvprPaperID{9888} 

\ifcvprfinal\fi
\begin{document}

\title{On Translation Invariance in CNNs: \\ Convolutional Layers can Exploit Absolute Spatial Location}

\author{Osman Semih Kayhan\\
Computer Vision Lab\\
Delft University of Technology\\
\and
Jan C. van Gemert\\
Computer Vision Lab\\
Delft University of Technology\\
}

\maketitle

\begin{abstract}

In this paper we challenge the common assumption that convolutional layers in modern CNNs are translation  invariant. We show that CNNs  can and will exploit the absolute spatial location by learning filters that respond exclusively to particular absolute locations by exploiting image boundary effects. Because modern CNNs filters have a huge receptive field, these boundary effects operate even far from the image boundary, allowing the network to exploit absolute spatial location all over the image. We give a simple solution to remove spatial location encoding which improves  translation invariance and thus gives a stronger visual inductive bias which particularly benefits small data sets. We broadly demonstrate these benefits on several architectures and various applications such as image classification, patch matching, and two video classification datasets.
\end{abstract}

\section{Introduction}

The  marriage of the convolution operator and deep learning yields the Convolutional Neural Network (CNN). The CNN arguably spawned the deep learning revolution with AlexNet~\cite{krizhevsky2012imagenet} and convolutional layers are now the standard backbone for various Computer Vision domains such as image classification~\cite{he2016deep, simonyan2014very, szegedy2015going}, object detection~\cite{liu2016ssd, redmon2016you, renNIPS15fasterRCNN}, semantic segmentation~\cite{ he2017maskRCNN, kirillov2019panoptic, ronneberger2015unet}, matching~\cite{longNIPS14convnetscorrespondence, han2015matchnet, zagoruyko2015compareImPatch}, video~\cite{carreira2017quo, hara2018can, simonyan2014two}, generative models~\cite{gatys2016image, goodfellow2014generative, kingma2018glow}, \etc. The CNN is now even used in other modalities such as speech~\cite{abdel2014convolutional, lecun1995convolutionalSpeech, oord2016wavenet}, audio~\cite{choi2017convolutionalMusic, hershey2017Audiocnn, salamon2017deepSound}, text~\cite{cho2014properties, dos2014deepText, lai2015recurrentText},  graphs~\cite{bruna2014spectral, duvenaud2015convolutionalGraph, schlichtkrull2018modeling}, \etc. It is difficult to overstate the importance of  the convolution operator in deep learning. In this paper we analyze  convolutional layers in CNNs which is broadly relevant for the entire deep learning research field. 

For images, adding convolution to neural networks adds a visual inductive prior that objects can appear anywhere. 
Convolution can informally be described as the dot product between the input image and a small patch of learnable weights --the kernel-- sliding over all image locations.  This shares the weights over locations yielding a huge reduction in learnable parameters. Convolution is equivariant to translation: If an object is shifted in an image then the convolution outcome is shifted equally. When convolution is followed by an operator that does not depend on the position, such as taking the global average or global maximum, that gives translation invariance and absolute location is lost. Translation invariance powers the visual inductive prior of the convolution operator, and we will demonstrate that improving translation invariance improves the prior, leading to increased data efficiency in the small data setting.

\begin{figure}
\centering
\begin{tabular}{cc}
Class 1: Top-left & Class 2: Bottom-right \\
\includegraphics[width=0.4\linewidth]{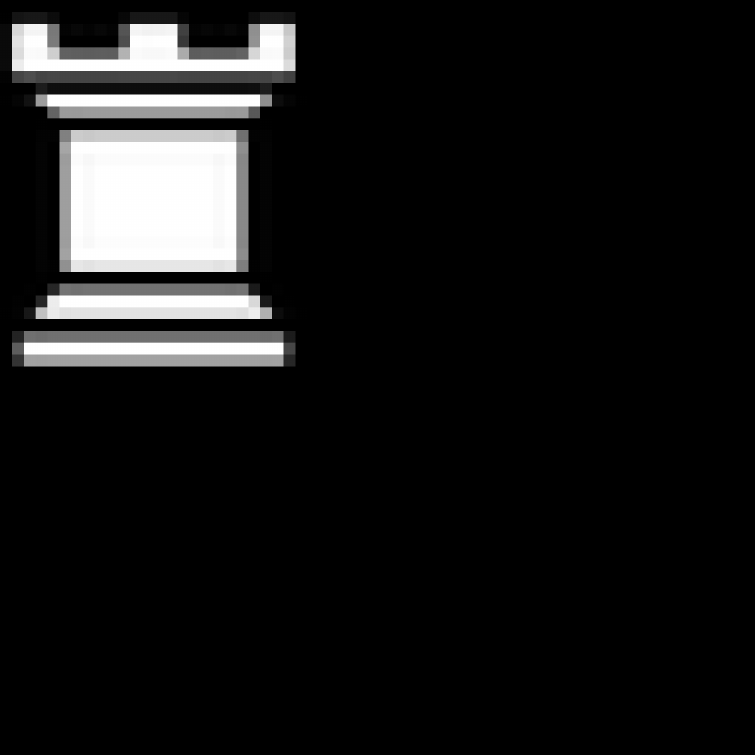} &
\includegraphics[width=0.4\linewidth]{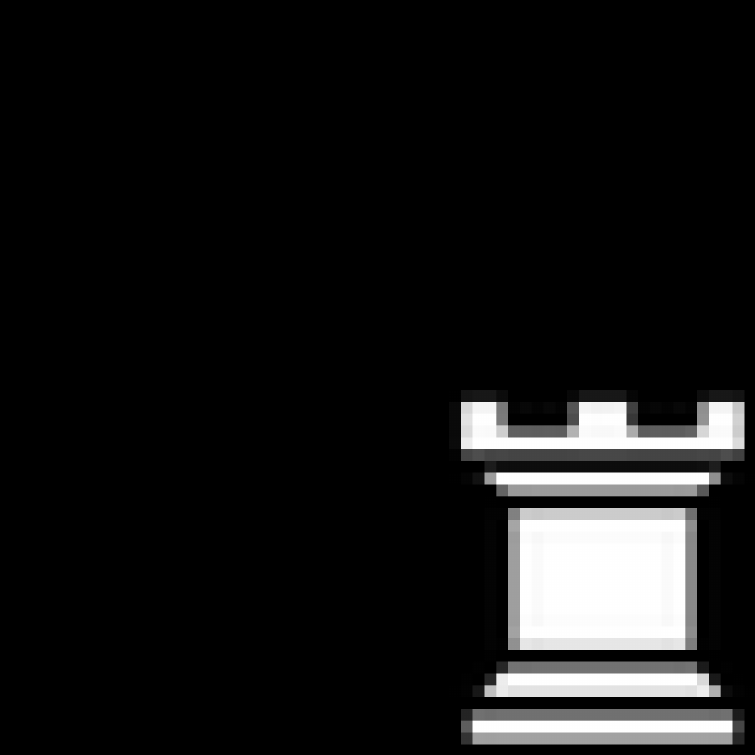}
\end{tabular}
\caption{We place an identical image patch on the top-left or on the bottom-right of an image. We evaluate a standard fully convolutional network~\cite{he2016deep, huang2017densely, lin2013NiN, springenberg2015striving, szegedy2015going, xieRexNext17CVPR} if it can classify the patch location (top-left vs bottom-right). We use 1 layer, a single 5x5 kernel, zero-padding, same-convolution,  ReLu, global max pooling, SGD, and a soft-max loss. Surprisingly, this network can classify perfectly, demonstrating that current  convolutional layers  can exploit the absolute spatial location in an image. }
\label{fig:quadrant}
\end{figure}

In this paper we challenge standard assumptions about translation invariance and  show that currently used convolutional layers can exploit the absolute location of an object in an image. Consider~\fig{quadrant}, where the exactly identical image patch is positioned on the top left (class 1) or on the bottom right (class 2) in an image.  If  a fully convolutional CNN is  invariant, it should not be able to classify and give random performance on this task. Yet, surprisingly, a simple standard 1-layer fully convolutional network with a global max pooling operator can perfectly classify the location of the patch and thus exploit absolute spatial location. 

We show that CNNs can encode absolute spatial location by exploiting image boundary effects. These effects occur because images have finite support and convolving close to the boundary requires dealing with non-existing values beyond the  image support~\cite{jahne2005digital, strang1996wavelets}. Boundary effects allow CNNs to learn filters whose output is placed outside the image conditioned on their absolute position in the image. This encodes position by only keeping filter outputs for specific absolute positions. It could, for example, learn filters that only fire for the top of the image, while the bottom responses are placed outside the image boundary. Boundary effects depend on the size of the convolution kernel and are small for a single 3x3 convolution. Yet, CNNs stack convolution layers, yielding receptive fields typically several times the input image size~\cite{araujoDISTILL19computingRF}. Boundary effects for such huge kernels are large and, as we will demonstrate, allows CNNs to exploit boundary effects all over the image, even far away from the image boundary. 

We have the following contributions. We show how boundary effects in discrete convolutions allow for location specific filters. We  demonstrate how convolutional layers in various current CNN architectures can and will exploit absolute spatial location, even far away from the image boundary. We investigate simple solutions that removes the possibility to encode spatial location which increases the visual inductive bias which is beneficial for  smaller datasets. We demonstrate these benefits on multiple CNN architectures on several application domains including image classification, patch matching, and video classification.

\section{Related Work and Relevance}

\textbf{Fully connected and fully convolutional networks}. Initial CNN variants have convolutional layers followed by fully connected layers. These fully connected layers can learn weights at each location in a feature map and thus can exploit  absolute position. Variants of the seminal  LeNet that included fully connected layers experimentally outperformed an exclusively convolutional setup~\cite{lecun1998gradient}. The 2012 ImageNet breakthrough as heralded by AlexNet~\cite{krizhevsky2012imagenet} followed the LeNet design, albeit at larger scale with 5 convolutional and 2 fully connected layers. Building upon AlexNet~\cite{krizhevsky2012imagenet}, the VGG~\cite{simonyan2014very} network family  variants involve varying the depth of the convolutional layers followed by 3 fully connected layers. The fully connected layers, however, take up a huge part of the learnable parameters making such networks large and difficult to train.

Instead of using fully connected layers, recent work questions their value. The Network In Network~\cite{lin2013NiN} is a fully convolutional network and simply replaces fully connected layers by the global average value of the last convolutional layer's output. Such a global average or global max operator is invariant to location, and makes the whole network theoretically insensitive to  absolute position by building on top of equivariant convolutional layers. Several modern networks are now using global average pooling. Popular and successful examples include the The All Convolutional Net~\cite{springenberg2015striving},  Residual networks~\cite{he2016deep}, The Inception family~\cite{szegedy2015going}, the DenseNet~\cite{huang2017densely}, the ResNext network~\cite{xieRexNext17CVPR} \etc. In this paper we show, contrary to popular belief, that  fully convolutional networks will exploit the absolute position.

\textbf{Cropping image regions.} Encoding absolute location has effect on cropping. Examples of region cropping in CNNs include: The bounding box in object detection~\cite{girshick2014rcnn, he2017maskRCNN, renNIPS15fasterRCNN}; processing a huge resolution image in patches~\cite{houCVPR16patch, sharma2017patch}; local image region matching~\cite{han2015matchnet, longNIPS14convnetscorrespondence, zbontar2016stereo, zagoruyko2015compareImPatch};  local CNN patch pooling encoders~\cite{arandjelovicCVPR16netvlad, babenko2015aggregating, brendelICLRbagnet}. The region cropping can be done \emph{explicitly} before feeding the patch to a CNN as done in R-CNN~\cite{girshick2014rcnn}, high-res image processing~\cite{houCVPR16patch} and aggregation methods~\cite{richardCVIU17bowRNN, shenCVPRw19facebagnet}. The other approach to cropping regions is \emph{implicitly} on featuremaps after feeding the full image to a CNN as done in 
Faster R-CNN~\cite{renNIPS15fasterRCNN}, 
BagNet~\cite{brendelICLRbagnet}, and CNN pooling methods such as sum~\cite{babenko2015aggregating}, BoW~\cite{passalis2017bowPooling}, VLAD~\cite{arandjelovicCVPR16netvlad, gong2014multi}, Fisher vector~\cite{cimpoi2016FVcnn}. In our paper we show that CNNs can encode the absolute position. This  means that in contrast to explicitly cropping a region before the CNN,  cropping a region after the CNN can include absolute position information, which impacts all implicit region cropping methods.

\textbf{Robustness to image transformations.} The semantic content of an image should be invariant to the accidental camera position. Robustness to such geometric transformation can be learned by adding  them to the training set using data augmentation~\cite{cubukCVPR19autoaugment, fawzi2016adaptive, hernandez2018dataAugment, hoICML19populationAugment, kauderer2017quantifyingTranslInv}. Instead of augmenting with random transformations there are geometric adverserial training methods~\cite{engstromICML19exploringSpatRobust,fawziBMVC15areReallyInv,KanbakCVPR18geometricRobustness} that intelligently add the most sensitive geometric transformations to the training data. Adding data to the training set by either data augmentation or  adverserial training is a brute-force solution   adding additional computation as the dataset grows. 

Instead of adding transformed versions of the training data there are methods specifically designed to learn geometric transformations in an equivariant or invariant
representation~\cite{biettiJMLR19groupInv, kondorICML18generalizationOfEquivar, LencCVPR15equivarAndEquival} where examples include rotation~\cite{dielemanICML16cyclicSymmetry, marcosICCV17rotationEqVectorField, weilerCVPR18steerableRotationEqv, worrallCVPR17harmonicNet, zhouCVPR17orientedResponseNets},
scale~\cite{marcos2018scale, sosnovik2019scale, wang2019selfSupScaleEq, worrall2019deepScale, xu2014scaleInvCNN} and other transformations~\cite{cohenICML16groupEquivariant, gensNIPS14deepSymmetry, henriquesICML17warpedConvs, laptevCVPR16transfInvPooling, sohn2012learningInvRepr}. 
Closely related is the observation that through subsequent pooling and subsampling in CNN layers  translation equivariance is lost~\cite{azulay2018CNNsmallTransform, zhangICML19blurPool}.  In our paper, we also investigate the loss of translation equivariance, yet do not focus on pooling but instead show that  convolutional layers can exploit image boundary effects to encode the absolute position which was also found independently by Islam~\etal~\cite{islam2019much}.


\textbf{Boundary effects.} Boundary effects cause statistical biases in finitely sampled data~\cite{griffith1983boundary, griffith1983evaluation}. For image processing this is textbook material~\cite{jahne2005digital, strang1996wavelets}, where boundary handling has applications in image restoration and deconvolutions~\cite{aghdasi1996reduction, liu2008reducing, reeves2005fast}. Boundary handling in CNNs focuses on minimizing boundary effects by learning separate filters at the boundary~\cite{innamorati2019learningBoundaryEffectsCNN}, treating out of boundary pixels as missing values~\cite{liu2018partialConvBasedPadding}, circular convolutions for wrap-around input data such as $360^\circ$ degree images~\cite{schubert2019circularForPanoramic} and minimizing distortions in $360^\circ$ degree video~\cite{cheng2018cubePadding}. We, instead, investigate how boundary effects can encode absolute spatial location.

\textbf{Location information in CNNs.} Several deep learning methods aim to exploit an absolute spatial location bias in the data~\cite{liuNIPS18coordConv, wang2018locationAugmentation}. This bias stems from how humans take pictures where for example a sofa tends to be located on the bottom of the image while the sky tends to be at the top. Explicitly adding absolute spatial location information helps for patch matching~\cite{luoECCV18geodescGeoConstraint, mukundan2019explicitSpatial}, generative modeling~\cite{wattersICLRW2019spatialBroadcast}, semantic segmentation~\cite{he2019locationawareSegmentation, wang2018locationAugmentation}, instance segmentation~\cite{nevenCVPR19instanceSpatial}. In this paper we do not add spatial location information. Instead, we do the opposite and show how to remove such absolute spatial location information from current CNNs.

\textbf{Visual inductive priors for data efficiency.} Adding visual inductive priors to deep learning increases data efficiency. Deep networks for image recognition benefit from a  convolutional prior~\cite{urbanICLR16CNNdeepAndC} and the architectural structure of a CNN with random weights already provides an inductive bias~\cite{jarrett2009bestMultiStageObjRecog, saxe2011random, ulyanov2018deepImPrior}. The seminial Scattering network~\cite{bruna2013scattering} and its variants~\cite{oyallon201rotoTranslationScattering, oyallon2018scattering} design a convolutional architecture to incorporate physical priors about image deformations. Other work shows that adding priors increases data efficiency by tying parameters~\cite{gensNIPS14deepSymmetry}, sharing rotation responses~\cite{worrallCVPR17harmonicNet}, and a prior scale-space filter basis~\cite{jacobsenCVPR16structuredRF}. In our paper we show that removing the ability of convolutional layers to exploit the absolute position improves translation equivariance and invariance which enforces the visual inductive prior of the convolution operator in deep learning.

\section{How boundary effects encode location}
\label{sec:method}

\begin{figure}
	\centering
	\includegraphics[width=0.95\linewidth]{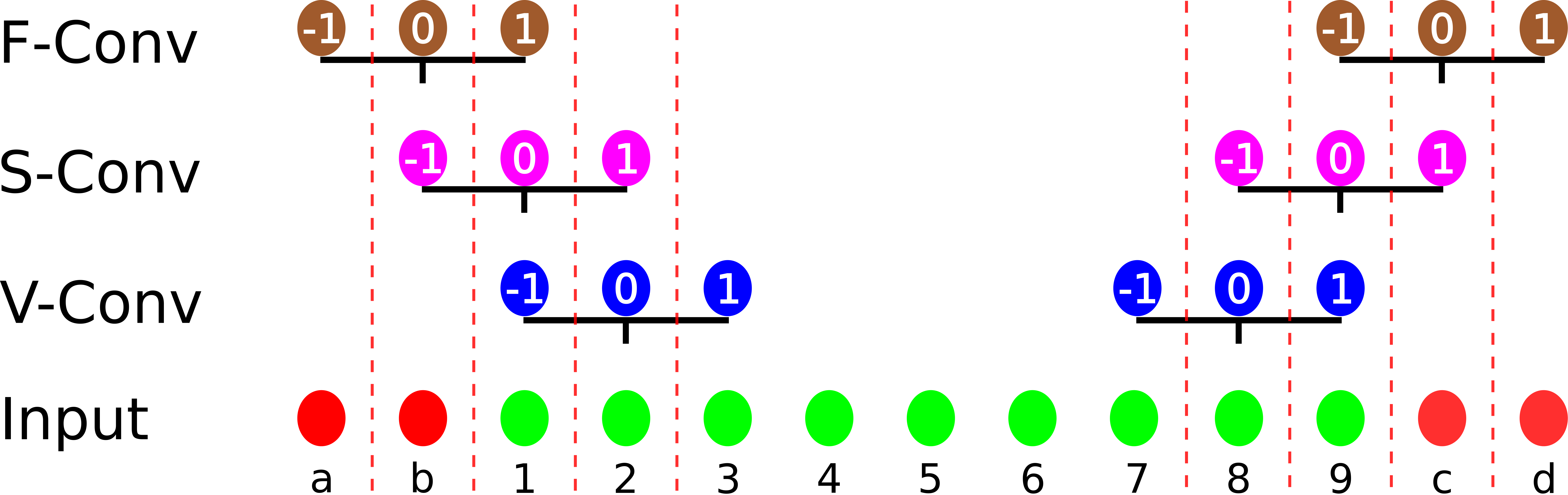}
	\caption{ How convolution ignores positions close to the border. We show the first and the last position for three convolution types: Valid (V-Conv), Same (S-Conv) and Full (F-Conv) applied to an input with finite support (green) and border padding (red). Note that for V-conv, the blue filter at position 1 is never applied to the green input positions 1 and 2. For S-Conv, the pink filter position 1 is never applied to green input position 1. F-Conv has all filter values applied on the image. }
	\label{fig:conv_border}
\end{figure}

We explore common convolution types for boundary handling with their image padding variants and explore their equivariant and invariant properties. In~\fig{conv_border} we illustrate the convolution types. For clarity of presentation we mostly focus on $d=1$ dimensional convolutions in a single channel, although the analysis readily extends to the multi-dimensional multi-channel case. We use the term 'image' broadly and also includes feature maps.

\textbf{Boundaries for convolution on finite samples.} 
Let ${\bf x} \in \mathbb{R}^n$ be the 1-D single channel input image of size~$n$ and ${\bf f} \in \mathbb{R}^{2k+1}$ denote a 1-D single channel filter where for convenience we only consider odd sized filters of size~$2k+1$. The output $y[t]$ for discrete convolution is
\begin{equation}
	{\bf y}[t] = \sum_{j=-k}^{k} {\bf f}[j] {\bf x}[t-j].
	\label{eq:finite_conv}
\end{equation}
Images have finite support and require handling boundary cases, for example where $t-j < 0$ and $x[t-j]$ falls outside the defined image. Providing values outside the image boundary is commonly referred to as padding. We consider two cases. \textit{Zero padding} assumes that all values outside of the images are zero. \textit{Circular padding} wraps the image values on one side around to the other side to provide the missing values.

\begin{figure*}
	\centering
	\includegraphics[width=\textwidth]{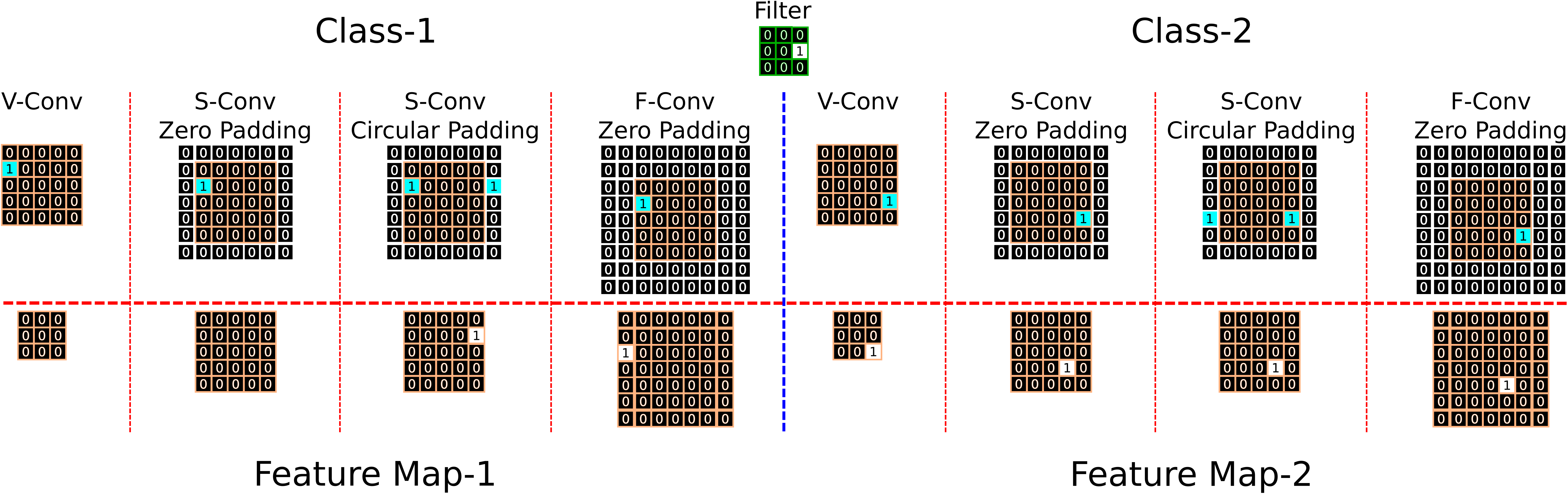}
	\caption{ A 2D Example where a pixel on the top left input (Class-1) and the same pixel on the bottom-right input (Class-2) can be classified using convolution. Comparing the output of 4 convolution types shows that V-Conv and S-Conv for Class-1 can no longer detect the pixel, while Class-2 still has the pixel. S-Conv with circular padding and F-Conv always retain the pixel value. }
	\label{fig:conv_methods}
\end{figure*}

\subsection{Common convolutions for boundary handling}

\textbf{Valid convolution (V-Conv).} V-Conv does not convolve across image boundaries. Thus, V-conv is a function $\mathbb{R}^n~\to~\mathbb{R}^{n-2k}$ where the output range of \eq{finite_conv} is in the interval:
\begin{equation}
t \in [k+1,n-k].
\label{eq:Vconv}
\end{equation}

It only considers existing values and requires no padding. Note that the support of the output $y$ has $2kd$ fewer elements than the input $x$, where $d$ is the dimensionality of the image, \ie, the output image shrinks with $k$ pixels at all boundaries. 

\textbf{Same convolution (S-Conv).} S-Conv slides only the filter center on all existing image values. The output range of \eq{finite_conv} is the same as the input domain; \ie the interval:
\begin{equation}
t \in [1,n].
\label{eq:Sconv}
\end{equation}
The support of the output $y$ is the same size as the support of the input $x$. Note that $2kd$ values fall outside the support of $x$, \ie,  at each boundary  there are $k$ padding values required.

\textbf{Full convolution (F-Conv).} F-Conv applies each value in the filter on all values in the image. Thus, F-conv is a function $\mathbb{R}^n~\to~\mathbb{R}^{n+2k}$ where the output range of \eq{finite_conv} is in the interval:
\begin{equation}
	t \in [-k,n+k].
\label{eq:Fconv}
\end{equation}
The output support of $y$ has $2kd$ more elements than the input $x$, \ie, the image grows with k elements at each boundary. Note that $4kd$ values fall outside of the support of the input $x$: At each boundary $2k$ padded values are required.

\subsection{Are all input locations equal?}

We investigate if convolution types are equally applied to all input position in an image.  In~\fig{conv_border} we illustrate the setting. To analyze if each location is equal, we modify~\eq{finite_conv} to count how often an absolute spatial position $a$ in the input signal $x$ is used in the convolution.  The count $C(\cdot)$ sums over all input positions $i$ where the convolution is applied,
\begin{equation}
C(a) = \sum_i  \sum_{j=-k}^k  \llbracket i = a-j \rrbracket,
\label{eq:countConv}
\end{equation}
where $\llbracket \cdot \rrbracket$ are Iverson~Brackets which evaluate to 1 if the expression in the brackets is true. Without boundary effects $C(a)$ always sums to $2k+1$ for each value of $a$.

\begin{figure*}
	\centering
	\begin{tabular}{c@{}c}
		\includegraphics[width=0.99\linewidth]{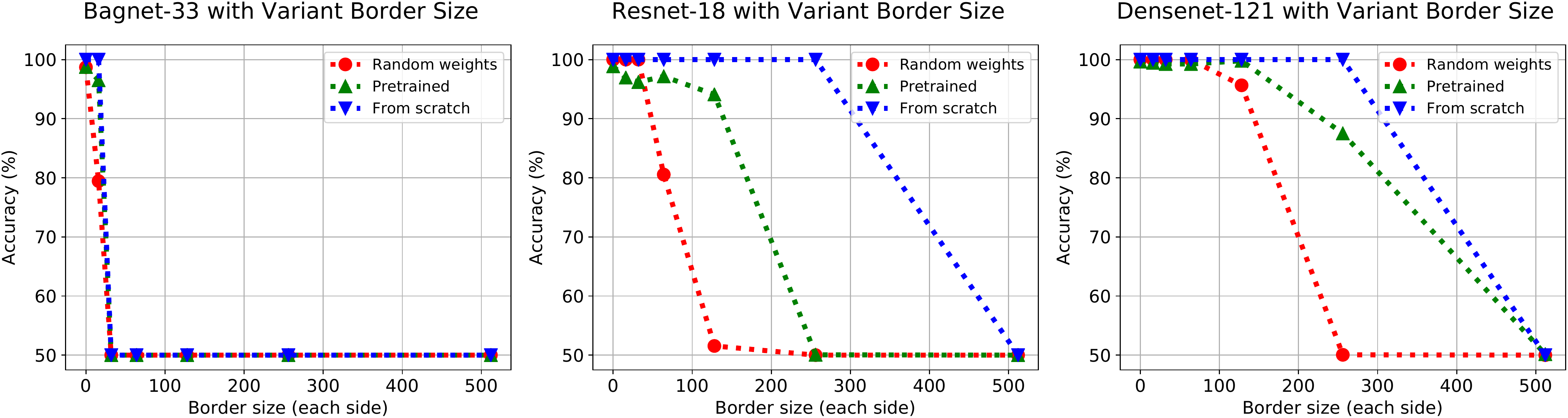}
	\end{tabular}
	\caption{ {\bf Exp 1:} Evaluating a BagNet-33~\cite{brendelICLRbagnet} (left), a ResNet-18~\cite{he2016deep} (middle) and a DenseNet-121~\cite{huang2017densely} (right) on how far from the boundary absolute location can be exploited, see~\fig{rf_block2}. The x-axis is the border size added to all 4 sides of the image and the y-axis is accuracy. All models can classify absolute position. The small RF of the BagNet allows for classification close to the border. The ResNet-18 and DenseNet-121 have larger RFs and can classify location far from the boundary. Random convolutional weights stay relatively close to the boundary while training on ImageNet learns filters that can go further. Training from scratch does best. Note that the most distant location from an image boundary for a $k$x$k$ image is a border size of $k/2$, \ie, a border size of 128 corresponds to a 256x256 image.}
	\label{fig:location_cls}
\end{figure*}

\begin{figure}
	\centering
	\includegraphics[width=0.99\linewidth]{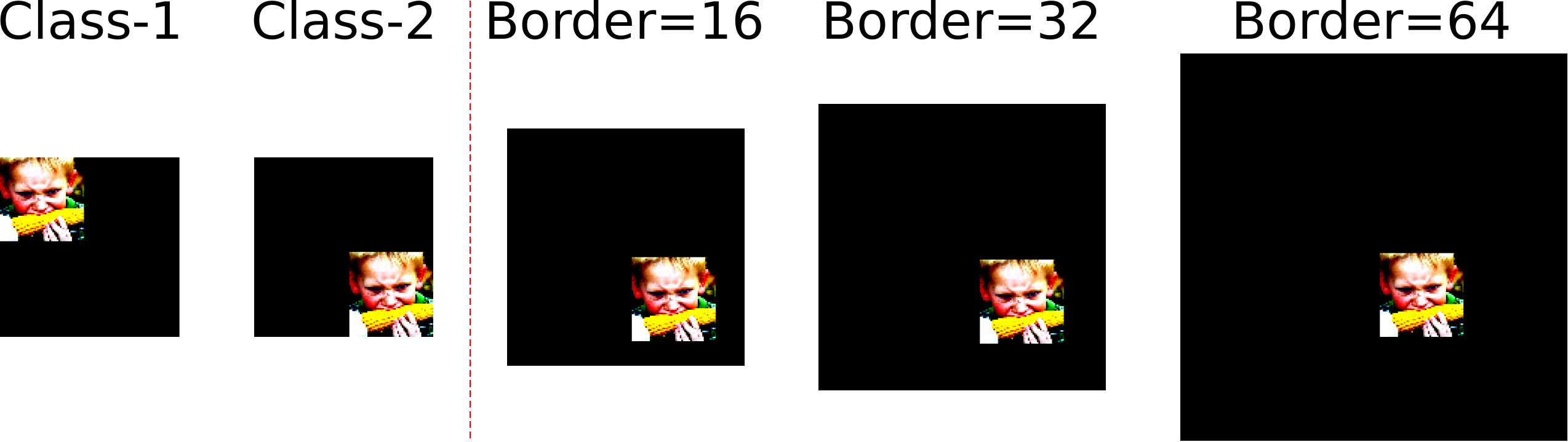}
	\caption{{\bf Exp 1:} Example images. Evaluating how far from the image boundary absolute location can be exploited. The task is to classify the location of a 56x56 resized Imagenet image placed in the top-left (class-1) and bottom-right (class-2), see also~\fig{quadrant}. We add a border on all 4 sides of the image, where we increase the border size until location can no longer be classified.}
	\label{fig:rf_block2}
\end{figure}

When there are boundary effects, there will be differences.
For V-Conv, the input locations $i$ are determined by \eq{Vconv} and the equation becomes
\begin{equation}
C_V(a) = \sum_{i=k+1}^{n-k} \sum_{j=-k}^k  \llbracket i = a-j \rrbracket,
\label{eq:countVConv}
\end{equation}
where $i$ no longer sums over all values. Thus, for all locations 
in the input image the function $C_V(t)$ no longer sums to $2k+1$ as it does in~\eq{countConv}, instead they sum to a lower value. In fact, it reduces to 
\begin{equation}
C_V(a) = 
\begin{cases}
a & \text{if }  a \in [1,2k] \\
n-a+1 & \text{if } a \in [n-2k, n]  \\
2k+1 & \text{Otherwise.}  
\end{cases}
\end{equation}
This shows that for V-Conv there are absolute spatial locations where the full filter is not applied.

For S-Conv, where~\eq{Sconv} defines the input, the count is
\begin{equation}
C_S(a) = \sum_{i=1}^{n} \sum_{j=-k}^k  \llbracket i = a-j \rrbracket,
\label{eq:countSConv}
\end{equation}
where $i$ sums over all values, and slides only the filter center over all locations. Thus, for S-Conv, when the locations are  $a \leq k$ or $a \geq n-k$, the function $C_S(a)$ no longer sums to $2k+1$. This reduces to 
\begin{equation}
C_S(a) = 
\begin{cases}
a+k & \text{if }  a \in [1,k] \\
n-a+(k+1) & \text{if } a \in [n-k, n]  \\
2k+1 & \text{Otherwise.}  
\end{cases}
\label{eq:casesSconv}
\end{equation}
This means that also for S-Conv there are absolute spatial locations where the full filter is not applied.

S-Conv with circular padding 'wraps around' the image and uses the values on one side of the image to pad the border on the other side. Thus, while for S-Conv,~\eq{casesSconv} holds for the absolute position $i$, it is by using circular padding that the \emph{value} $x[i]$ at position $i$ is exactly wrapped around to the positions where the filter values were not applied. Hence, circular padding equalizes all responses, albeit at the other side of the image. Zero padding, in contrast, will have absolute spatial locations where filter values are never applied.

For F-Conv, in~\eq{Fconv}, the counting equation becomes 
\begin{equation}
C_F(a) =  \sum_{i=-k}^{n+k} \sum_{j=-k}^k  \llbracket i = a-j \rrbracket.
\label{eq:countFConv}
\end{equation}
F-Conv sums the filter indices over all indices in the image and thus, as in~\eq{countConv}, all locations $i$ sum to $2k+1$ and thus no locations are left out.

We conclude that V-Conv is the most sensitive to exploitation of the absolute spatial location. S-Conv with zero padding is also sensitive to location exploitation. S-Conv with circular padding is not sensitive, yet involves wrapping values around to the other side, which may introduce semantic artifacts. F-Conv is not sensitive to location information. In~\fig{conv_methods} we give an example of all convolution types and how they can learn absolute spatial position.

\section{Experiments}

\textbf{Implementation details for Full Convolution.} For standard CNNs implementing F-Conv is trivially achieved by simply changing the padding size. For networks with residual connections, we add additional zero padding to the residual output to match the spatial size of the feature map. We will make all our experiments and \href{https://github.com/oskyhn/CNNs-Without-Borders}{code} available\footnote{\url{https://github.com/oskyhn/CNNs-Without-Borders}}.

\subsection{Exp 1: How far from the image boundary can absolute location be exploited? }

CNNs can encode absolute position by exploiting boundary effects. In this experiment we investigate how far from the boundary these effects can occur. Can absolute position be encoded only close to the boundary or also far away from the boundary? To answer this question we revisit the location classification setting in~\fig{quadrant} while adding an increasingly large border all around the image until location can no longer be classified. In~\fig{rf_block2} we show the setting.


We randomly pick 3,000 samples from ImageNet validation set, resize them to 56x56 and distribute them equally in a train/val/test set. For each of the 3k samples we create two new images (so, 2,000 images in each of the 3 train/val/test sets) by taking a black 112x112 image and placing the resized ImageNet sample in the top-left corner (class-1) and in  the bottom-right corner (class-2), see~\fig{quadrant}. 
To evaluate the distance from the boundary we create 7 versions by adding a black border of size  $\in \{0, 16, 32, 64, 128, 256, 512\}$ on all 4 sides of the 112x112 image, see~\fig{rf_block2} for examples.





We evaluate three networks with varying receptive field size. BagNet-33~\cite{brendelICLRbagnet} is a ResNet variant where the receptive field is constrained to be 33x33 pixels. ResNet-18~\cite{he2016deep} is a medium sized network, while a DenseNet-121~\cite{huang2017densely} is slightly larger. We evaluate three settings: (i) trained completely from scratch to see how well it can do; (ii) randomly initialized  with frozen convolution weights to evaluate the architectural bias for location classification; (iii) ImageNet pre-trained with frozen convolution weights to evaluate the location classification capacity of a converged realistic  model used in a typical image classification setting. 

Results in~\fig{location_cls} show that all settings for BagNet, ResNet and DenseNet can classify absolute position. Random weights can do it for locations relatively close to the boundary. Surprisingly, the pre-trained models have learned filters on ImageNet that can classify position further away from the boundary as compared to random initialization. The models trained from scratch can classify absolute position the furthest away from the boundary. The BagNet fails for locations far from the boundary. Yet, the medium-sized ResNet-18 can still classify locations of 128 pixels away from the boundary, which fully captures ImageNet as for 224x224 images the most distant pixel is only 112 pixels from a boundary. We conclude that absolute location can even be exploited far from the boundary.

\subsection{Exp 2: Border handling variants}

\begin{figure}
\centering
    \includegraphics[width=0.99\linewidth]{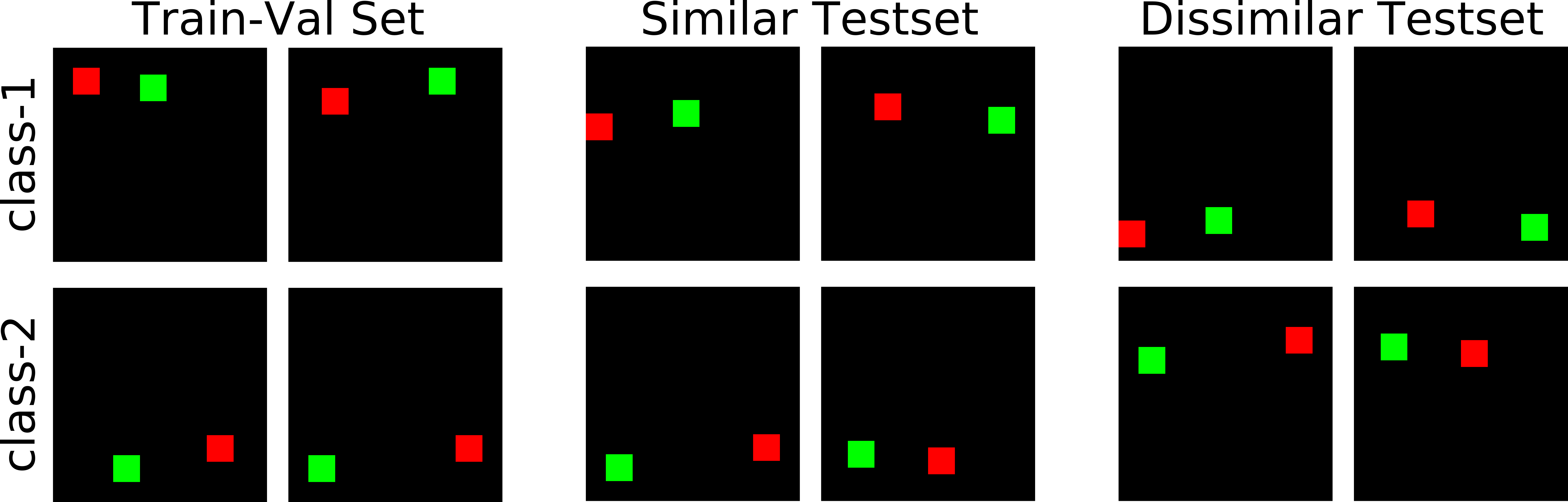}
\caption{{\bf Exp 2:} Example images of the Red-Green two class classification dataset for evaluating exploitation of absolute position. The upper row of images is class 1: Red-to-the-left-of-Green. The lower row of images is class 2: Green-to-the-left-of-Red. The Similar Testset is matching the Train-Val set in absolute location: Class 1 at the top and class 2 at the bottom. The Dissimilar testset is an exact copy of the Similar testset where absolute location is swapped between classes: Class 1 at the bottom, Class 2 at the top.  If absolute location plays no role then classification on the Similar Testset would perform equal to the Dissimilar Testset.  }
\label{fig:red_green_all}
\end{figure}

Border handling is the key to absolute location coding. Here we evaluate the effect of various border handling variants on absolute location exploitation. To do so, we create an image classification task unrelated to the absolute position and introduce a location bias which should have no effect on translation invariant architectures.

We construct the Red-Green data set for binary image classification of the relative order of colored blocks of 4x4 on a black 32x32 image. Class 1 has Red to the left of Green; class 2 has Green to the left of Red, see~\fig{red_green_all}. The classification task is unrelated to the absolute position. We introduce a vertical absolute position bias by placing class 1 on the top of the image (8 pixels from the top, on average), and class 2 on the bottom (8 pixels from the bottom, on average). We then construct two test sets, one with similar absolute location bias, and a dissimilar test set where the location bias switched: class 1 at the bottom and class 2 on top, see~\fig{red_green_all}. 

The train set has 2,000 images, the validation and test sets each have 1,000 images. Experiments are repeated 10 times with different initialization of the networks. A 4-layer fully convolutional deep network is used for evaluation. The first two layers have 32 filters and last two layers 64 filter followed by global max pooling. Sub-sampling for layers 2, 3, 4 uses stride 2 convolution. 

We evaluate the border handling of \sect{method}. \emph{V-Conv} uses only existing image values and no padding. For \emph{S-Conv} we evaluate zero and circular padding. \emph{F-Conv} has zero padding. Results are in~\tab{red_green_results}. \emph{V-Conv} and \emph{S-Conv}-zero have the best accuracy on the Similar test set, yet they exploit the absolute location bias and perform poorly on the Dissimilar test set, where \emph{V-Conv} relies exclusively on location and confuses the classes completely. \emph{S-Conv}-circ and \emph{F-Conv} perform identical on the Similar and Dissimilar test sets; they are translation invariant and thus cannot exploit the absolute location bias. \emph{F-Conv} does better than \emph{S-Conv}-circ because circular padding introduces new content. \emph{F-Conv} does best on both test sets as it is translation invariant and does not introduce semantic artifacts.

\begin{table}
	\centering
	\begin{tabular}{lccc}  \toprule
		Type & Pad  & Similar test   & Dissimilar test \\ \midrule
		V-Conv & - & $\bf{100.0} \pm 0.0 $ & $0.2 \pm 0.1$ \\
		S-Conv & Zero & $99.8  \pm 0.1$ & $8.4  \pm  0.7$ \\
		S-Conv  & Circ & $73.7 \pm 1.0 $ & $73.7 \pm 1.0$ \\
		F-Conv  & Zero & $89.7 \pm 0.5$  & $\bf{89.7} \pm 0.5$ \\ 
		\bottomrule
	\end{tabular}
	\caption{{\bf Exp 2:} Accuracy on the Red-Green dataset shown in \fig{red_green_all}. Type is the convolution type, pad is how padding is done. Results are given on the Similar test set with matching absolute positions and the Dissimilar test set with an absolute position mismatch. Stddevs are computed by 10 repeats. \emph{Valid} and \emph{same}-zero exploit location and do poorly on the Dissimilar test set. \emph{Same}-circ is translation invariant yet invents disturbing new content. \emph{Full}-zero is translation invariant, doing well on both test sets. }
	\label{tab:red_green_results}
\end{table}

\begin{figure*}
	\centering
	\begin{tabular}{c@{\hspace{1mm}}c}
		\includegraphics[width=0.99\linewidth]{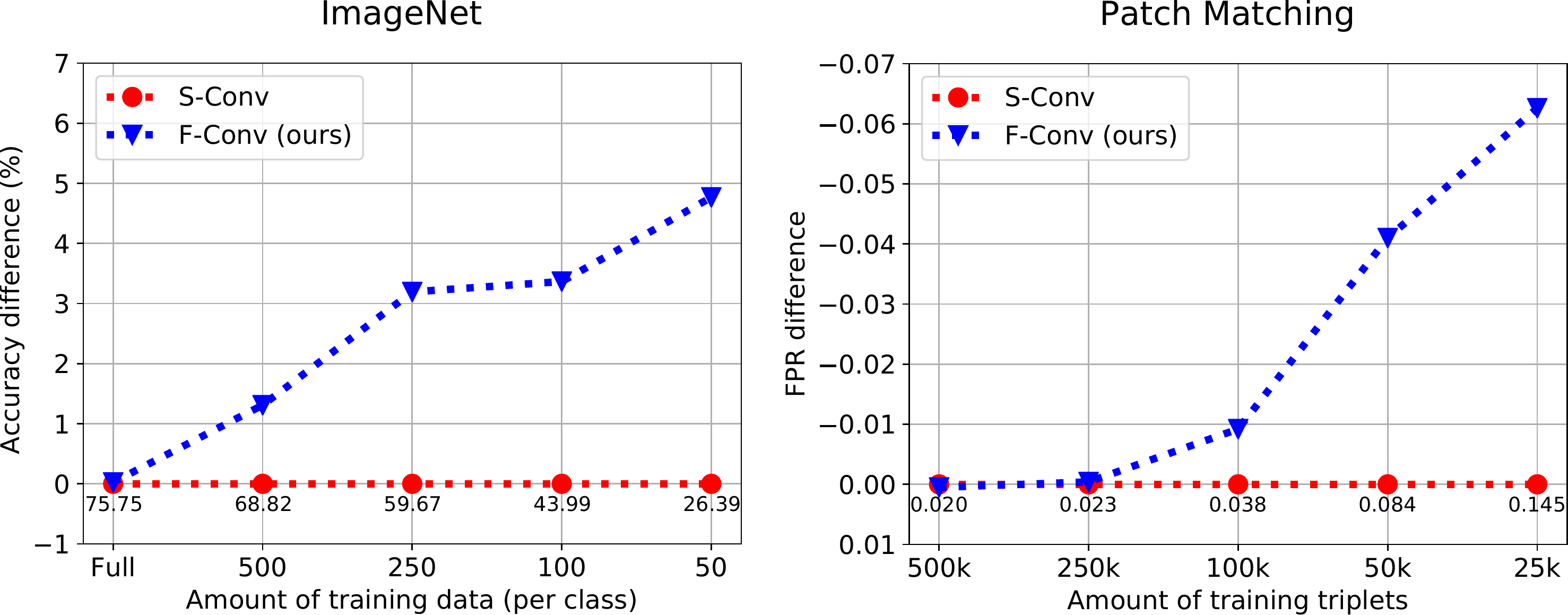} 
	\end{tabular}
	\caption{{\bf Exp 4:} Data efficiency experiments. We reduce the amount of training data per class for the 1,000 classes of Imagenet for image classification and full Liberty, Notre Dame and Yosemite for patch matching. F-Conv outperforms S-Conv in both modality with smaller data size. (left) The Imagenet plot demonstrates the obtained accuracy difference when the number of data samples per class. The difference between F-Conv and S-Conv increases when the sample size decreases. (right) Correspondingly, F-Conv results in a performance increase for patch matching. }
	\label{fig:reducing_tra}
\end{figure*}

\subsection{Exp 3: Sensitivity to image shifts}


Does removing absolute location as a feature lead to robustness to location shifts? We investigate the effect of image shifts at test time on CNN output for various architectures on a subset of ImageNet.  We train four different architectures from scratch with S-Conv and F-Conv: Resnet~18,~34,~50 and 101. To speed up training from scratch, we use 20\% of the full ImageNet and take the 200 classes from~\cite{hendrycks2019natural} which is still large but 5x faster to train. To evaluate image shifts we follow the setting of BlurPool~\cite{zhangICML19blurPool}, which investigates the effect of pooling on CNN translation equivariance. As BlurPool improves equivariance, we also evaluate the effect of BlurPool Tri-3~\cite{zhangICML19blurPool}.

\begin{table}[]
	\centering
	\resizebox{\linewidth}{!}{
		\begin{tabular}{lcccc}  \toprule
			Diagonal\\ Shift & S-Conv & F-Conv & S+BlurPool & F+BlurPool \\ \midrule
			RN18                                                     & 79.43 & 82.74  &  81.96          &    \textbf{83.95}       \\
			RN34                                                     & 82.06 & 85.66  &  83.73          &     \textbf{86.91}      \\
			RN50                                                     & 86.36 & 87.92  &  87.50       &       \textbf{88.93}     \\
			RN101                                                    & 86.95 & 87.78  &  88.22          &    \textbf{88.73}  \\ \toprule
			Consistency                                              & S-Conv & F-Conv & S+BlurPool & F+BlurPool \\ \midrule
			RN18                                                     & 86.43 & 88.38 &   88.32         &    \textbf{90.03}        \\
			RN34                                                     & 87.62 & 90.12 &   89.21         &     \textbf{91.53}       \\
			RN50                                                     & 90.21 & 91.36 &   91.68       &     \textbf{92.75}        \\
			RN101                                                    & 90.76 & 91.71 &   92.36       &      \textbf{92.86}       \\           
		\bottomrule		
		\end{tabular}%
}
\caption{{\bf Exp 3:} Diagonal shift and consistency result for different Resnet architectures. S+BlurPool represents S-Convs with BlurPool Tri-3. Similarly, F+BlurPool corresponds the combination of F-Conv and BlurPool. In the most cases, F-Conv outperforms S-Conv and S+BlurPool (except for Resnet-101) in terms of  diagonal shifting accuracy on testing set. Similar trend can be seen for consistency experiment, yet for Resnet-50 and Resnet-101, S+BlurPool has more consistent outputs. F+BlurPool achieves the highest score for both cases with all the architectures.}
\label{tab:image_shift}
\end{table}

\textbf{Diagonal Shift.} We train the network with the usual central crop. Each testing image is diagonally shifted starting from the top-left corner towards the bottom-right corner. We shift 64 times 1 pixel diagonally. Accuracy is evaluated for each pixel shift and averaged over the full test set.

\textbf{Consistency.} We measure how often the classification output of a model is the same for a pair of randomly chosen diagonal shifts between 1 and 64 pixels~\cite{zhangICML19blurPool}. We evaluate each test image 5 times and average the results.

Results are given in~\tab{image_shift}.
For each architecture, using F-Conv improves both the classification performance and the consistency of all the models. The highest classification accuracy gain between S-Conv and F-Conv is 3.6\% and the best consistency gain is 2.49\% with Resnet-34.  BlurPool makes S-Convs more robust to diagonal shifts and increase consistency. When F-Conv and BlurPool are combined, the accuracy on diagonal shifting and consistency are improved further. Resnet-34 (F+BlurPool) obtains more 4.85\% of accuracy and 3.91\% of consistency compared to the S-Conv baseline. If we compare each Resnet architecture, the deepest model of the experiment, Resnet-101, improves the least, both for classification and consistency. Resnet-101 has more filters and parameters and it can learn many more varied filters than other models. By this, it can capture many variants of location of objects and thus the gap between methods for Resnet-101 are smaller. 

\subsection{Exp 4: Data efficiency}

Does improving equivariance and invariance for the inductive convolutional prior lead to benefits for smaller data sets? We evaluate S-Conv and F-Conv with the same random initialization seed for two different settings: Image classification and image patch matching.

\textbf{Image classification.} We evaluate ResNet-50 classification accuracy for various training set sizes of the 1,000 classes in ImageNet.  We vary the training set size as 50, 100, 250, 500, and all images per class.

\textbf{Patch matching.}
We use HardNet~\cite{mishchuk2017hardNet} and use FPR (false positive rate) at 0.95 true positive recall as an evaluation metric (lower is better). We evaluate on 3 common patch matching datasets (Liberty, Notre Dame and Yosemite) from Brown dataset~\cite{Brown2007} where the model is trained on one set and tested on the other two sets. Hardnet uses triplets loss and we vary the training set size as 50k, 100k, 250k, 500k triplet patches. Each test set has 100k triplet patches. 


Results are given in~\fig{reducing_tra}. For both image classification as for patch matching S-Conv and F-Conv perform similar for a large amount of training data. Yet, when reducing the number of training samples there is a clear improvement for F-Conv. For ImageNet with only 50 samples per class S-Conv scores 26.4\% and F-Conv scores 31.1\%, which is a relative improvement of 17.8\%. For patch matching, S-Conv scores 0.145 and F-Conv 0.083 which is a relative improvement of 75\%. Clearly, removing absolute location improves data efficiency.


%


%

\subsection{Exp 5: Small datasets}
%



\begin{table}[]
	\centering
	\resizebox{\linewidth}{!}{%
		\begin{tabular}{@{}lcccc@{}}
			\toprule
			& \multicolumn{2}{c}{UCF101}                                                                                            &  \multicolumn{2}{c}{HMDB51}                                                                                            \\  \cmidrule(r){2-3} \cmidrule(l){4-5} 
			& \begin{tabular}[c]{@{}c@{}}Baseline\\ (S-Conv)\end{tabular} & \begin{tabular}[c]{@{}c@{}}Ours\\ (F-Conv)\end{tabular} & \begin{tabular}[c]{@{}c@{}}Baseline\\ (S-Conv)\end{tabular} & \begin{tabular}[c]{@{}c@{}}Ours\\ (F-Conv)\end{tabular} \\
			RN-18 & 38.6                                                        & 40.6                                                    & 16.1                                                        & 19.3                                                    \\
			RN-34 & 37.0                                                        & 46.9                                                    & 15.2                                                        & 18.3                                                    \\
			RN-50 & 36.2                                                        & 44.1                                                    & 14.3                                                        & 19.0                                                    \\ \bottomrule
		\end{tabular}%
	}
	\caption{{\bf Exp 5:} Action recognition with 3D Resnet-18, 34 and 50 by using S-Conv and F-Conv methods. F-Conv outperforms S-Conv on UCF101 and HMDB51 datasets. S-Conv obtains its best result with the most shallow network, Resnet-18, however F-Conv still improves the results even the model becomes bigger. }
	\label{tab:action}
\end{table}

Here we evaluate if the improved data efficiency generalize to two small datasets for action recognition. We select small sized data sets where training from scratch gives significantly worse results due to overfitting and the common practice is pre-training on a huge third party dataset. We compare the standard S-Conv with the proposed F-Conv where both methods are trained from scratch.

\textbf{Action Recognition.}
We evaluate on two datasets:  UCF101~\cite{DBLP:journals/corr/abs-1212-0402} with 13k video clips from 101 action classes and HMDB51~\cite{kuehne2011hmdb} with 51 action classes and around 7k annotated video clips. We evaluate three 3D Resnet architectures~\cite{hara2018can}, Resnet-18, 34 and 50.

\begin{figure}
	\centering
	\begin{tabular}{c@{}c}
		\includegraphics[width=0.49\linewidth]{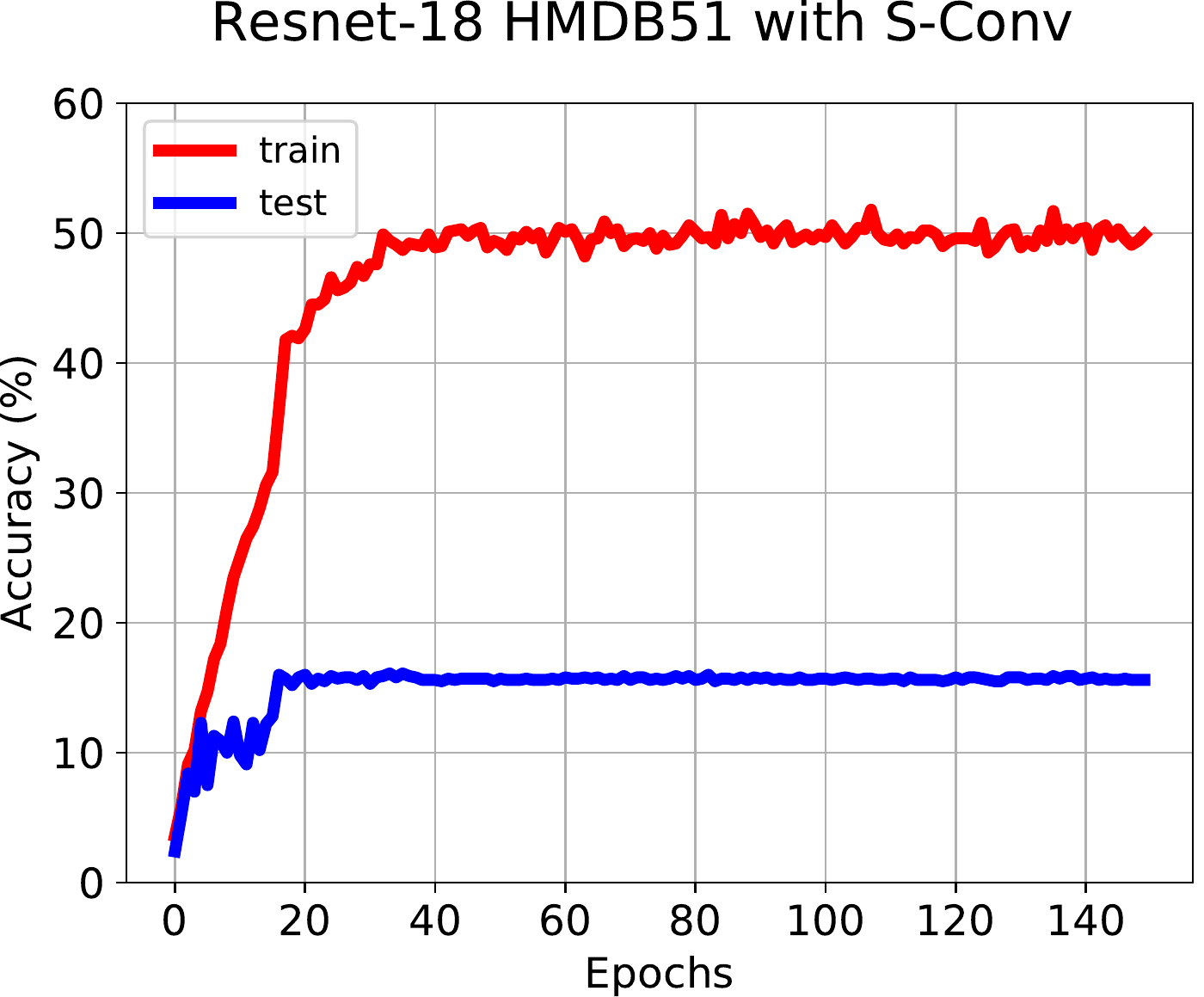} &
		\includegraphics[width=0.49\linewidth]{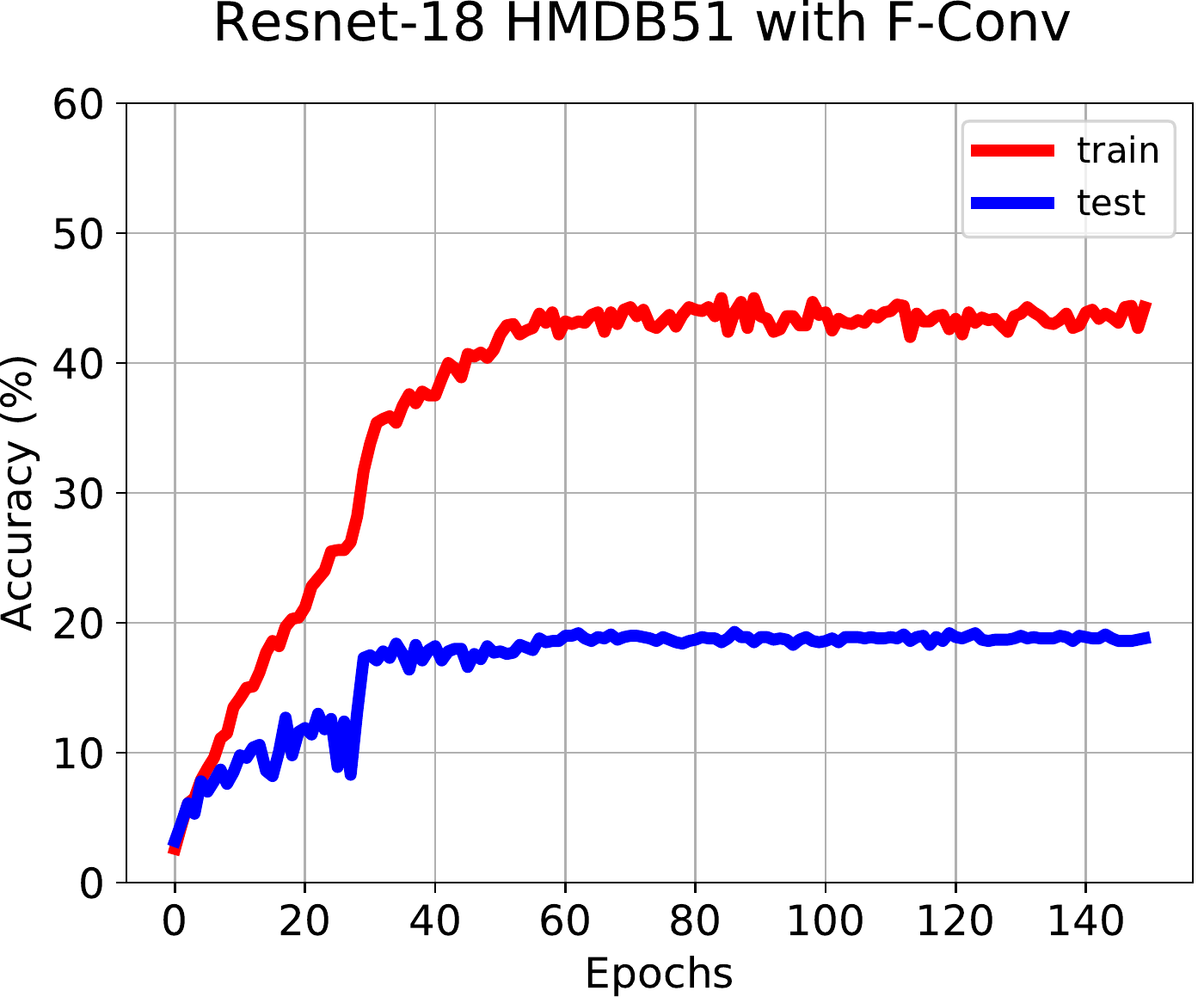} 
	\end{tabular}
	\caption{{\bf Exp 5:} Training curves for 3D Resnet-18 S-Conv (left) and F-Conv (right) with HMDB51 dataset. Because the dataset is small, both models overfit. F-Conv achieves relatively 38.8\% less overfitting than S-Conv.}
	\label{fig:HMDB51}
\end{figure}
\begin{figure}
	\centering
	\begin{tabular}{c@{}c}
		\includegraphics[width=0.49\linewidth]{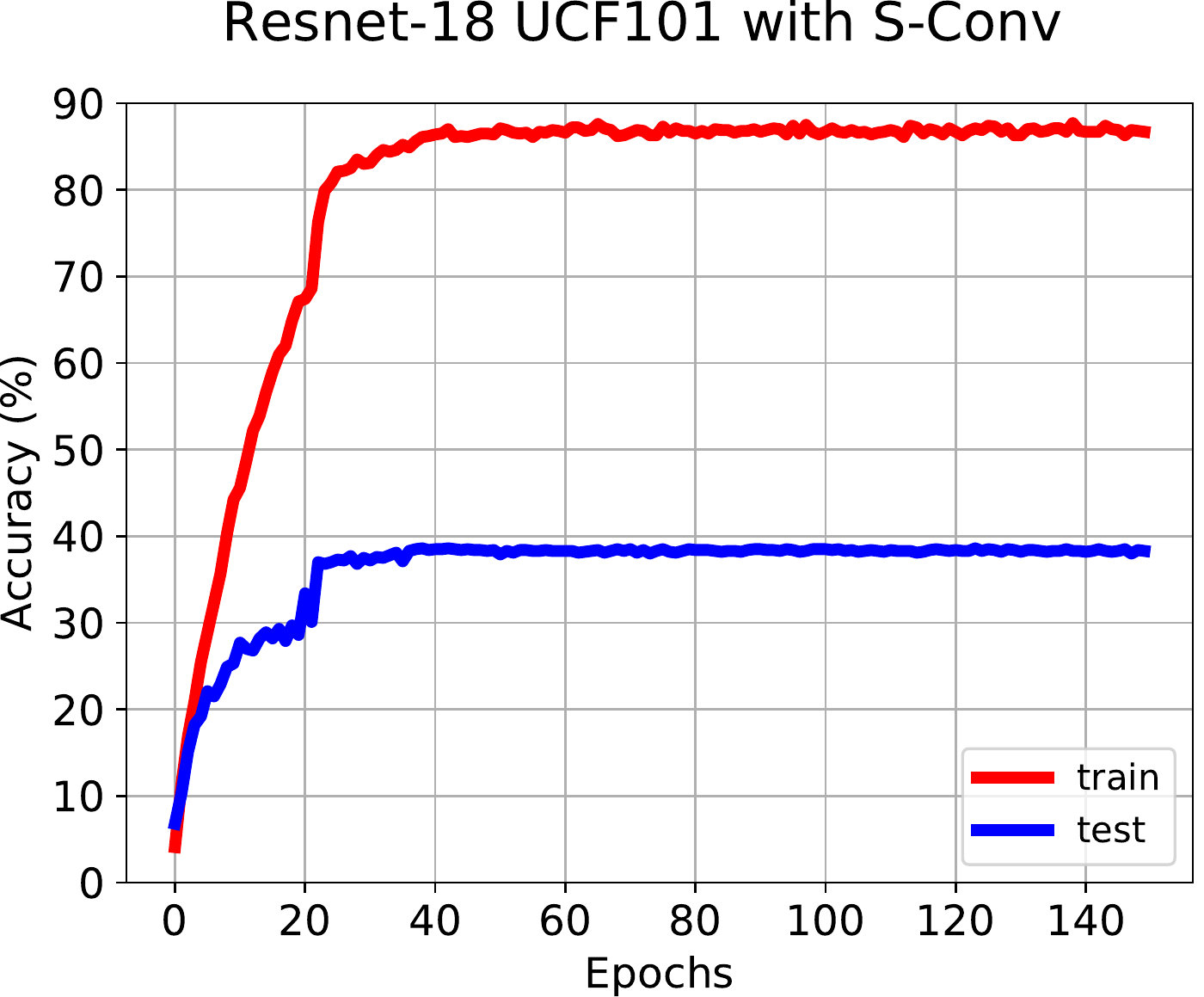} &
		\includegraphics[width=0.49\linewidth]{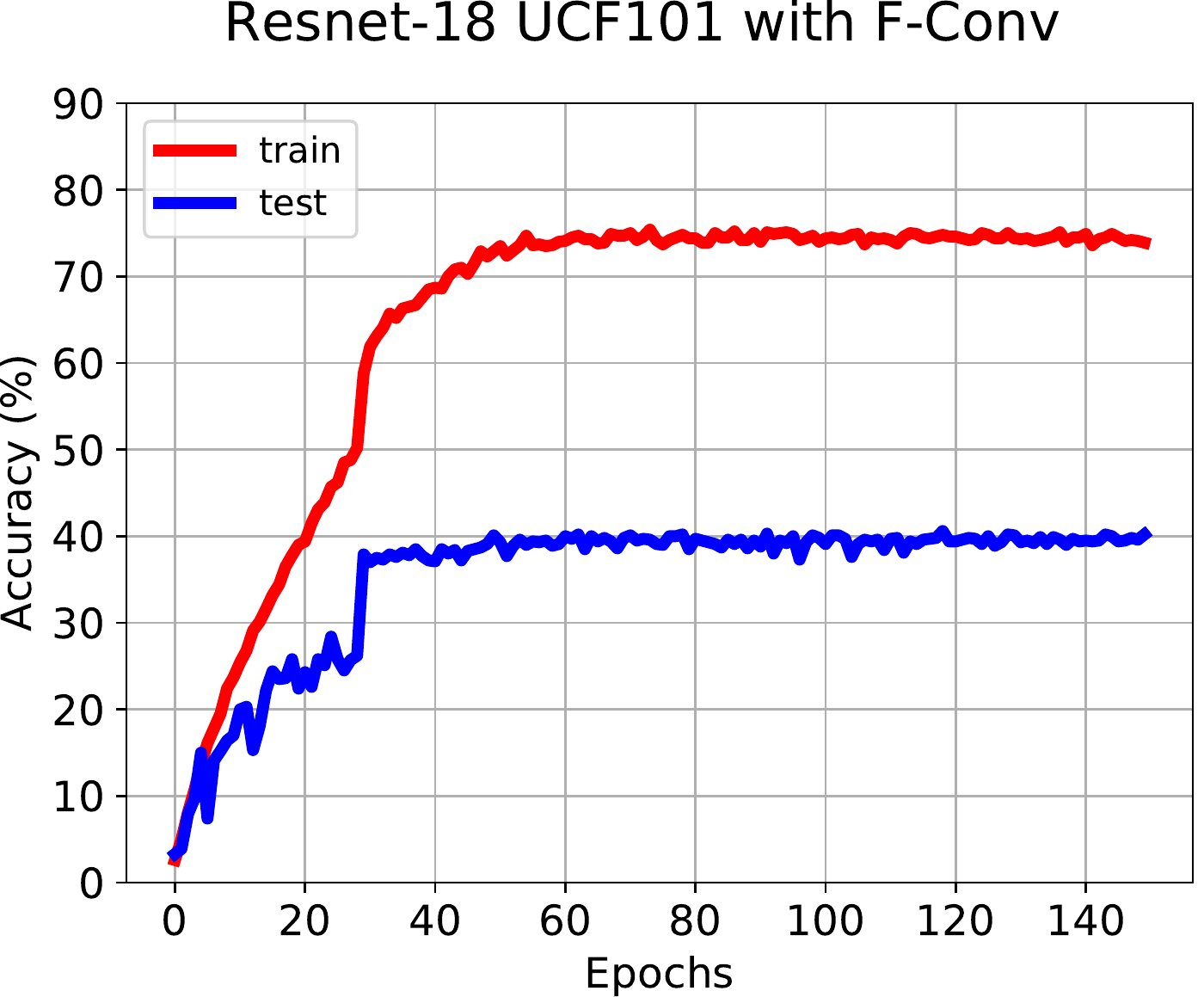} 
	\end{tabular}
	\caption{{\bf Exp 5:} Training curves for 3D Resnet-18 S-Conv (left) and F-Conv (right) with UCF101 dataset. Both models overfit, but S-Conv has higher difference between training and testing results (49.1\%). F-Conv has 34.8\% of gap and thus overfits less. }
	\label{fig:UCF101}
\end{figure}

We show results in~\tab{action}. F-Conv models outperform the S-Conv models. Interestingly, in UCF101 experiment, the baseline performance decreased by 2.4\% from Resnet-18 to Resnet-50; however, F-Convs still continue to improve the performance by 3.6 \% for same architectures. According to Kensho et al ~\cite{hara2018can} a 3D Resnet-18 overfits with UCF101 and HMDB51 which we confirm, yet F-Conv we overfit less than S-Conv. In~\fig{HMDB51}, the difference between train and test of a 3D Resnet18 with S-Conv is 35.69\%, however F-Conv has 25.7\% overfitting. Similarly, S-Conv is relatively 41\% more overfitted than F-Conv in~\fig{UCF101}. Consequently, both methods overfit due to the number of parameter and the lack of data.

\section{Limitations and Conclusion}

One limitation of our method is the extra computation required for padding.  There is no extra cost of using circular padding instead of zero padding. For using F-Conv instead of S-Conv, the costs are similar to using S-Conv instead of V-Conv, and we found a Resnet-50 with F-Conv 15\% slower to train on Imagenet.

Note that if absolute spatial location is truly discriminative between classes, it \emph{should} be exploited~\cite{van2011exploiting}, and not removed. For many internet images with a human photographer, there will be a location bias as humans tend to take pictures with the subject in the center, sofas on the bottom, and the sky up. The difficulty lies in having deep networks not exploit spurious location correlations due to lack of data. Addressing lack of data samples by sharing parameters over locations through added convolutions in deep networks is a wonderfully regularizer and we believe that convolutional layers should truly be translation equivariant.

To conclude, we show that in contrary to popular belief, convolutional layers can encode the absolute spatial location in an image. With the strong presence of the convolution operator in deep learning this insight is relevant to a broad audience. We analyzed how boundary effects allow for ignoring certain parts of the image. We evaluated existing networks and demonstrated that their large receptive field makes absolute spatial location coding available all over the image. We demonstrate that removing spatial location as a feature increases the stability to image shifts and improves the visual inductive prior of the convolution operator which leads to increased accuracy in the low-data regime and small datasets which we demonstrate for ImageNet image classification, image patch matching, and two video classification data sets.

{\small
\bibliographystyle{ieee_fullname}
\bibliography{egbib}
}

\end{document}